\documentclass[times,referee,twocolumn,final,authoryear]{elsarticle}

\usepackage{ycviu}
\usepackage{framed,multirow}

\usepackage{amssymb}
\usepackage{latexsym}

\usepackage{url}
\usepackage{xcolor}
\definecolor{newcolor}{rgb}{.8,.349,.1}

\usepackage{textcomp}
\usepackage{multirow}
\usepackage{amsmath}
\usepackage{graphicx}
\PassOptionsToPackage{normalem}{ulem}
\usepackage{ulem}
\usepackage{amsfonts}
\usepackage{algorithmic}
\usepackage{algorithm}
\usepackage{array}
\usepackage{stfloats}
\usepackage{url}
\usepackage{verbatim}
\usepackage{pifont}
\newcommand{\cmark}{\ding{51}}
\newcommand{\xmark}{\ding{55}}

\journal{Computer Vision and Image Understanding}

\begin{document}

\ifpreprint
  \setcounter{page}{1}
\else
  \setcounter{page}{1}
\fi

\begin{frontmatter}

\title{Self-training via Metric Learning for Source-Free Domain Adaptation of Semantic Segmentation}

\author[1]{Ibrahim Batuhan \snm{Akkaya}\corref{cor1}} 
\cortext[cor1]{Corresponding author:}
\ead{bthakkaya@gmail.com}
\author[1]{Ugur \snm{Halici}}

\address[1]{Middle East Technical University, Ankara 06800, Turkey}

\received{1 May 2013}
\finalform{10 May 2013}
\accepted{13 May 2013}
\availableonline{15 May 2013}
\communicated{S. Sarkar}

\begin{abstract}
  Unsupervised source-free domain adaptation methods aim to train a model for the target domain utilizing a pretrained source-domain model and unlabeled target-domain data, particularly when accessibility to source data is restricted due to intellectual property or privacy concerns. Traditional methods usually use self-training with pseudo-labeling, which is often subjected to thresholding based on prediction confidence. However, such thresholding limits the effectiveness of self-training due to insufficient supervision. This issue becomes more severe in a source-free setting, where supervision comes solely from the predictions of the pre-trained source model. In this study, we propose a novel approach by incorporating a mean-teacher model, wherein the student network is trained using all predictions from the teacher network. Instead of employing thresholding on predictions, we introduce a method to weight the gradients calculated from pseudo-labels based on the reliability of the teacher's predictions. To assess reliability, we introduce a novel approach using proxy-based metric learning. Our method is evaluated in synthetic-to-real and cross-city scenarios, demonstrating superior performance compared to existing state-of-the-art methods.

\end{abstract}

\begin{keyword}
\MSC 41A05\sep 41A10\sep 65D05\sep 65D17
\KWD Keyword1\sep Keyword2\sep Keyword3

\end{keyword}

\end{frontmatter}


\section{Introduction}

Recent advancements in deep learning have led to notable improvement in the field of computer vision. Deep learning networks trained in a supervised manner demonstrate high performance, even in challenging tasks that require dense prediction, such as semantic segmentation \citep{chen2017rethinking, chen2017deeplab, wang2020deep, zhong2020squeeze, xu2023pidnet, fan2021rethinking}. However, the preparation of large datasets with dense labeling is a laborious and expensive process \citep{cordts2016cityscapes}. One approach to mitigate the workforce requirements involves the use of publicly available real datasets \citep{everingham2010pascal, cordts2016cityscapes, chen2017no} or synthetic datasets \citep{Richter_2016_ECCV, Ros_2016_CVPR}. However, it is important to note that deep learning approaches typically assume the Identically and Independently Distributed (IID) distribution of training and testing sets. Given the distinct marginal distributions between the source and target domains, the trained model experiences performance degradation when evaluated on the target data — a phenomenon known as the domain gap.

\begin{figure}[t]
\centering

\includegraphics[width=1\columnwidth]{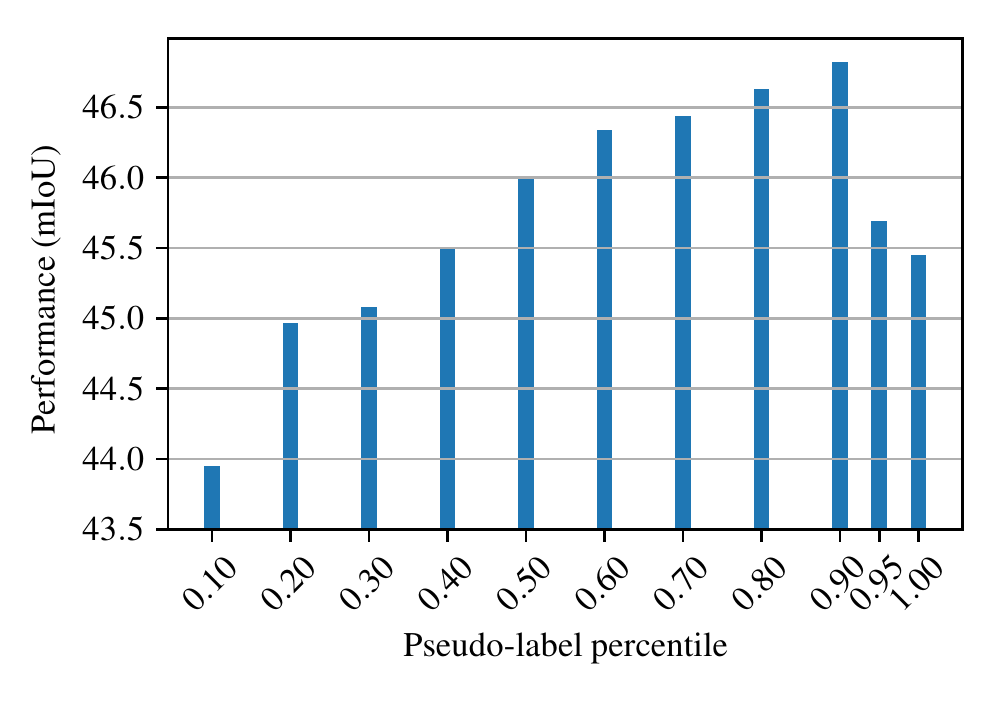}

\caption{Segmentation performance of the self-training when different percentile
of the predictions are used as pseudo-labels}

\label{fig:st_quantile_results}
\end{figure}

Various domain adaptation methods have been introduced to address the challenge of domain gap \citep{sankaranarayanan2018learning,du2019ssf,tsai2018learning,araslanov2021self,pan2020unsupervised,xu2021neutral,zhou2020affinity, hoyer2023mic}. These approaches leverage both labeled data from a source domain and unlabeled data from a target domain during model training. However, in certain applications such as medical imaging or autonomous driving, accessing labeled source domain data may be restricted due to intellectual property (IP) or privacy concerns. To overcome this limitation, an alternative solution involves utilizing a model trained on the source domain instead of the actual data. This approach, known as source-free domain adaptation, aims to reduce the domain gap by employing a source model alongside unlabeled target domain data. 

Self-training is a widely utilized technique in both classical domain adaptation and source-free domain adaptation \citep{liu2021cycle, herath2023energy, karim2023c}. This training strategy involves adapting the model using pseudo-labels generated by the model itself. Many approaches incorporate prediction filtering mechanisms to determine these pseudo-labels \citep{zou2018unsupervised,zou2019confidence,mei2020instance,li2019bidirectional}. In the context of source-free domain adaptation, where supervision relies exclusively on the target dataset due to the absence of a source dataset, harnessing predictions from target images becomes paramount during training. Thresholding the predictions in this scenario leads to a degradation of supervision. Conversely, utilizing all predictions is suboptimal, as erroneous predictions can misguide the training process, resulting in performance deterioration, as illustrated in Figure \ref{fig:st_quantile_results}. Consequently, instead of employing thresholding, our approach seeks to utilize all predictions by scaling the guidence based on the reliability of each prediction. 

Assessing the reliability of predictions within the target domain presents a formidable challenge, as neural networks often exhibit overconfident false predictions when encountering data significantly deviating from learned training set patterns, such as out-of-distribution images \citep{hein2019relu}. In the context of domain adaptation, where target dataset images are inherently out-of-distribution for a model trained on the source domain, establishing a domain-specific and reliable metric becomes imperative. Addressing this challenge, we propose a novel Self-Training via Metric learning (STvM) reliability measure tailored to the target domain. 
Our method uses pixel-level proxy-based metric learning to train a network that predicts distance metrics for each pixel in target domain images. The goal is to keep pixels from the same class close together, while keeping them far from pixels of other classes. By using proxy-based metric learning, we simultaneously train a feature vector for each class. These vectors act as class prototypes in the target domain. We then compute the reliability measure by evaluating the distance between the pixel feature and the proxy feature of the predicted class.
This methodology greatly improves the model's ability to assess reliability in the target domain. This is especially important when dealing with out-of-distribution images, a common situation given that images from the target domain are considered out-of-domain for a model trained in the source domain. Furthermore, we leverage the reliability metric to sample patches from the target domain, allowing us to adapt mixing augmentation effectively in the absence of labeled data.

In summary, our contributions are as follows:

\begin{itemize}
\item We introduce a novel reliability metric, learned directly in the target domain through a proxy-based metric learning approach.
\item Unlike conventional methods that filter predictions for pseudo-labels, we employ all predictions in the self-training process and adjust the gradients based on the reliability metric.
\item We devise an effective way to adapt mixing augmentation, dubbed as a metric-based online ClassMix, in the absence of labeled data by leveraging the reliability metric to sample patches from the target domain.
\item Our STvM (Self-Training via Metric-learning) approach demonstrates superior performance compared to state-of-the-art methods across GTA5-to-CityScapes, SYNTHIA-to-CityScapes, and NTHU datasets.
\end{itemize}


\section{Related works}

\paragraph{Domain adaptation}

Domain adaptation is a widely studied topic, especially for image classification \citep{wang2018deep} and semantic segmentation \citep{toldo2020unsupervised}. Adversarial learning \citep{chang2019all,hong2018conditional,tsai2018learning,chen2019learning,chen2019crdoco,choi2019self,li2019bidirectional} and self-training \citep{chen2019domain,iqbal2020mlsl,mei2020instance,pan2020unsupervised,zhang2019category,zheng2021rectifying,zou2019confidence,zou2018unsupervised} approaches are commonly utilized in semantic segmentation. Adversarial methods, inspired by the GAN framework \citep{goodfellow2014generative}, align the feature space of the source and target. Unlike the GAN framework, these methods distinguish between source and target domains rather than real and fake samples. Adversarial learning is applied in image and feature levels. Some methods exploit the adversarial approach to the low-dimensional output space to facilitate adversarial learning. Self-training methods utilize confident predictions as pseudo-labels. These are obtained by filtering out noisy samples or applying constraints, and then used for supervised training.

\paragraph{Source-free domain adaptation}

Source-free domain adaptation applies domain adaptation using only the trained source model instead of the source data as well as unlabeled
target data, where the source data cannot be freely shared due to Intellectual Property of privacy concerns. The task is introduced by two works \citep{fleuret2021uncertainty,liu2021source} concurrently. URMA \citep{fleuret2021uncertainty} reduces the uncertainty of predictions in the presence of feature noise. This is accomplished by using multiple decoders and incorporating feature noise through dropout. The model's resistance to noise is preserved by an uncertainty loss, determined by the squared difference between the decoder outputs. The training stability is improved with entropy minimization and self-training with threshold-based pseudo-labeling. SFDA \citep{liu2021source} introduces a data-free knowledge distillation approach. It generates source domain synthetic images that preserve semantic information using batch-norm statistics of the model and dual-attention mechanism. They also use a self-supervision module to improve performance. 
HCL \citep{huang2021model} proposed Historical Contrastive Learning (HCL), comprising Historical Contrastive Instance Discrimination (HCID) and Historical Contrastive Category Discrimination (HCCD). HCID contrasts embeddings of target samples from current and historical models, while HCCD employs pseudo-labels for learning category-discriminative representations.
LD \citep{you2021domain} highlights that self-training methods often face the 'winner-takes-all' issue, where majority classes dominate, leading to segmentation networks failing to classify minority classes. To remedy this, they propose a two-component framework: positive and negative learning. Positive learning selects class-balanced pseudo-labeled pixels using an intra-class threshold, while negative learning identifies the category to which each pixel does not belong through complementary label selection.
DA+AC \citep{yang2022source} presents a framework to stabilize SFDA for semantic segmentation. Initially, in the distribution transfer phase, it aligns source model features with target data statistics. Then, in the self-training stage, it employs adaptive thresholding to select per-class pseudo labels for self-supervision.
GtA \citep{kundu2021generalize} categorized the SF-UDA methods as a vendor and client-side strategies, where vendor-side strategies focus on the improving source model for better adaptation. They focus on improving the vendor side performance and propose multiple augmentation techniques to train different source models using the leave-one-out technique on the vendor side.

Considering that it may not be possible to train the model on the vendor-side in source-free scenario, we propose a client-side source free domain adaptation method like SFDA and URMA, in which the source model is used as-is.

\paragraph{Metric learning}

Deep metric learning (DML) is an approach to establish a distance metric between data to measure similarity. It learns an embedding space where similar examples are close to each other, and different examples are far away. It is a widely studied topic in computer vision that has various applications such as image retrieval \citep{movshovitz2017no}, clustering \citep{hershey2016deep}, person re-identification \citep{cheng2016person} and face recognition \citep{schroff2015facenet}. In the unsupervised domain adaptation literature, metric learning is mostly utilized in the image classification task. The features of the semantically similar samples from both source and target domain are aligned to mitigate the domain gap \citep{huang2015cross,laradji2020m,pinheiro2018unsupervised,yu2018correcting}. Commonly, the similarity metric between images or patches is trained in deep metric learning. With a different perspective, Chen et al. \citep{chen2018blazingly} utilize pixel-wise deep metric learning for interactive object segmentation, where they model the task as an image retrieval problem. DML approaches are categorized into two classes based on the loss functions, namely pair-based and proxy-based methods. While pair-based loss functions \citep{wang2019multi,wang2019ranked,oh2016deep,sohn2016improved} exploit the data-to-data relations, the proxy-based loss functions \citep{movshovitz2017no,teh2020proxynca++,qian2019softtriple,aziere2019ensemble} exploit data-to-proxy relations. Generally, the number of proxies is substantially smaller than training data. Therefore, proxy-based methods converge faster, and the training complexity is smaller than pair-based methods. They are also more robust to label noise and outliers \citep{kim2020proxy}. 

In STvM, we propose a proxy-based pixel-wise metric learning to estimate pseudo-label reliability, trained by limited and noisy pseudo-labels. 

\paragraph{Mixing}

Mixing is an augmentation technique that combines pixels of two training images to create highly perturbed samples. It has been utilized in classification \citep{berthelot2019mixmatch,yun2019cutmix,zhang2018mixup} and semantic segmentation \citep{french2019semi,olsson2021classmix}. Especially in semi-supervised learning of semantic segmentation, mixing methods, such as CutMix \citep{yun2019cutmix} and ClassMix \citep{olsson2021classmix}, achieved promising results. While the former cut rectangular regions from one image and paste them onto another, the latter use a binary mask belonging to some classes to cut. DACS \citep{tranheden2021dacs} adapts ClassMix \citep{olsson2021classmix} for domain adaptation by mixing across domains.

In source-free domain adaptation, access to the source domain dataset is not available. This makes it impossible to combine target images with source patches. Additionally, the absence of labels in the target domain dataset presents a challenge when trying to use ClassMix. In order to utilize mixing, we propose an metric based online ClassMix method  where patches are sampled based on our reliability metric.

\section{Proposed method}

In this section, we present our self-training via metric learning (STvM) approach for the unsupervised source-free domain adaptation problem in the context of semantic segmentation. The source-free domain adaptation setting is composed of the following components. Let $\mathcal{D}_{S}$ is a labeled source dataset composed of source domain images $x_{S}$ and corresponding pixel-wise labels $y_{S}$, $\mathcal{D}_{S}=\{(x_{S}^{i},y_{S}^{i})\}_{i=1}^{N_{S}}$ where $N_{S}$ is the number of samples in $\mathcal{D}_{S}$. It is assumed that the source model is trained with $\mathcal{D}_{S}$ in a supervised manner so that it performs well in the source domain. Let $\mathcal{D}_{T}$ is an unlabeled target dataset composed of target domain images, $\mathcal{D}_{T}=\{(x_{T}^{i})\}_{i=1}^{N_{T}}$ where $N_{T}$ is the number of samples in $\mathcal{D}_{T}$. The source-free domain adaptation technique involves training a model using a pretrained model from the source domain and an unlabeled dataset from the target domain $\mathcal{D}_{T}$. Our approach leverages the mean teacher model, which comprises two segmentation networks: the teacher network $\mathcal{T}$ and the student network $\mathcal{S}$. We aim to incorporate source domain knowledge during the initial training phase and gradually adapt to the target domain. To achieve this, we initialize both models with the source domain-trained model. The student network's parameters are updated using  gradient descent on the target dataset $\mathcal{D}_{T}$ through a self-training approach. Meanwhile, the teacher network's parameters are updated by exponentially averaging the student network's parameters throughout the training process, facilitating a smooth transition from the source to the target domain.

\subsection{Framework overview}

\begin{figure*}[t]
\centering
\includegraphics{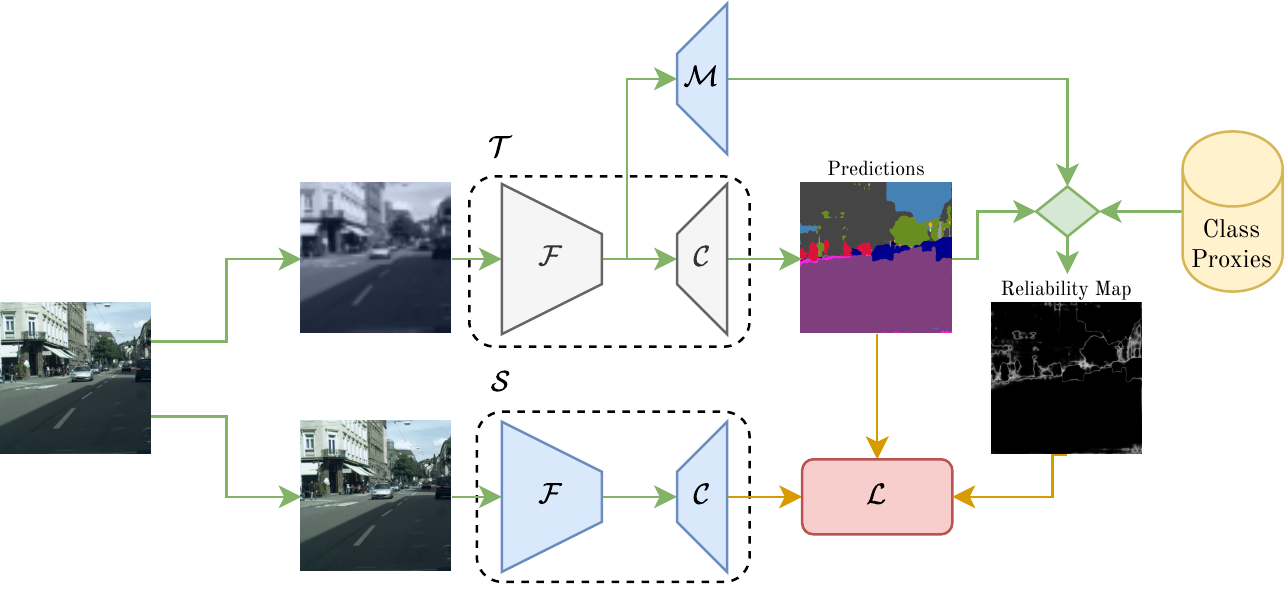}
\caption{STvM comprises three networks, namely teacher, student, and metric network, represented as $\mathcal{T}$, $\mathcal{S}$, and $\mathcal{M}$, respectively. The teacher and the student network use the same segmentation network architecture. Each segmentation network is composed of feature extractor $\mathcal{F}$ and classifier $\mathcal{C}$. The metric network has the same architecture as $\mathcal{C}$, trained to learn metric feature space. Inspired by the mean-teacher approach, the student network is trained with a backpropagation algorithm. On the other hand, the parameters of the teacher model are updated with the moving average of the parameters of the student model. The metric network $\mathcal{M}$ and class proxies are also trained with a backpropagation algorithm.}
\label{fig:overview}
\end{figure*}

The proposed method is illustrated in the Figure \ref{fig:overview}. We utilize the mean-teacher approach to train a segmentation model.  There are two data paths in STvM: One belongs to the teacher and metric networks, and the other belongs to the student. 

The teacher data path is responsible for training the metric network, generating the pseudo-labels and the reliability map. An image belonging to the target dataset is fed to the teacher network. The predictions of the teacher model are used as pseudo-labels. In addition to that, the features generated by the feature extractor of the teacher network are fed to the metric network to form metric features for each pixel of the input image. Class proxy features, trained in conjunction with the metric network, are vectors that represent the class distribution in the metric space. The distance between the proxy feature of the predicted class and the feature generated by the metric network is utilized as a measure of class similarity. This class similarity is subsequently transformed into a reliability score using the reverse sigmoid function. The metric network is trained with a proxy-based metric learning approach using the highly confident predictions of the teacher network.

The student data path is responsible for the training of the student network using pseudo-labels and reliability scores. Firstly, A photometric augmentation is applied to the input image, then Metric-based Online ClassMix (MOCM) augmentation is applied to the input image, the pseudo-labels, and the reliability map. The patches that are used in MOCM are stored to the patch buffer in training time based on the metric similarity. The online patch update approach enables end-to-end training of the student model. The student model is trained with a pixel-wise cross-entropy loss function. However, the pixel-wise loss is multiplied with a reliability score to scale gradients which suppress erroneous pseudo-labels and allow under-confident accurate predictions to contribute to training. 

\subsection{Mean teacher with noisy student}

The mean teacher method constitutes two networks called the teacher and the  tudent networks \citep{tarvainen2017mean}. The parameters of the teacher network are updated by moving average of the student network parameters where the student network is updated with backpropagation. The mean teacher approach can be integrated with the self-training concept by generating the pseudo-labels from the teacher network and training the student network with these pseudo-labels. In order to obtain pseudo-labels as accurately as possible, we do not apply augmentation to the input of the student network. The continuous update of the teacher network enables gradually improved pseudo-labels during training. Xie et al. \citep{xie2020self} showed that deliberately adding noise to the student model leads to a better teacher model. Following the same principle, we applied an input noise using photometric augmentation and MOCM.

For the segmentation task, the teacher network $\mathcal{T}$ takes an image $x_{T}^{i}\in X_{T}\subset\mathbb{R}^{H\times W\times3}$ and generate a class probability distribution $p_{T}\in P_{T}\subset\mathbb{R}^{H\times W\times C}$ for each pixel, where $H$, $W$, and $C$ correspond to the image height, the image width and the number of classes, respectively. The pseudo-label map $\tilde{y}\in\mathbb{R}^{H\times W\times C}$ is a one-hot vector for each pixel, where the channel corresponding to the maximum prediction confidence is one and the others are zero.

The pixel-wise cross-entropy loss is a widely used criterion in semantic segmentation tasks. We adapted the cross-entropy loss to train the student network. Different from the existing methods, all pseudo-labels are taken into account to train the student network. However, the gradients computed by pseudo-labels are scaled based on the label's reliability. We multiply pixel-wise loss values with the gradient scaling factor using Hadamard multiplication to scale the gradients. The gradient scaling factor $w^{(w,h)}$ is a real-valued map taking values between 0 and 1, and it is calculated based on the reliability metric, explained in section \ref{subsec:Reliability-metric} in detail.

\subsection{Reliability metric learning\label{subsec:Reliability-metric}}

The gradient scaling factor ($w$) plays an important role in the training of the student model. It has a strong effect on the parameters of the student model since it manipulates the gradients. In addition to that, the student model parameters directly affect the teacher model's performance and consequently the quality of the pseudo-labels. Therefore, It is essential to estimate a good reliability metric in the target domain to calculate the scaling factor. To this end, we proposed a novel reliability metric predicted by the metric network $\mathcal{M}$. The metric network utilizes the same architecture with the classifier $\mathcal{C}$ of the segmentation network. It takes the features of the feature extractor network $\mathcal{F}$ of the teacher network $\mathcal{T}$, and predicts a pixel-wise metric feature $f\in\mathbb{R}^{H\times W\times N_{f}}$, corresponding one metric feature for each input image pixel. The metric network $\mathcal{M}$ learns a transformation from the segmentation feature space to the metric feature space, where the features of the pixels belonging to the same class are pulled together, and those of the pixels belonging to different classes are pushed away. 

The metric learning methods commonly use two different relations to train the model, namely pair-based and proxy-based relations. The pair-based methods use data-to-data similarity to train the network, whereas the proxy-based network uses proxy-to-data similarity. We
exploit a proxy based relation in our study. The proxy feature $f_{p}\in\mathbb{R}^{N_{f}}$
is a vector that represents the distribution of a class of data points.
We assign a different proxy feature to each segmentation class since
we want to estimate how much the data point is associated with the
predicted class. The proxy features are defined as a trainable parameter
just as the metric learning network parameters, and it is trained
concurrently with the metric network parameters. 

Metric learning is a supervised learning method that requires class
labels to be trained. Since the proxy-based method is known to be
trained with a small number of samples \citep{movshovitz2017no,teh2020proxynca++},
we use a small subset of the predictions of the teacher network with
high confidence values to train the metric network. This subset is
called the metric pseudo-labels, and they are selected with class
balanced thresholding strategy. 

\begin{equation}
\tilde{y}_{M}^{h,w,c}=\begin{cases}
\begin{array}{c}
1\\
0
\end{array} & \begin{array}{c}
p_{T}^{h,w,c}=\max(p_{T}^{h,w})>\tau_{c}\\
elsewhere
\end{array}\end{cases}.\label{eq:metric-pseudo-label}
\end{equation}

Threshold values ($\tau$) are varied for each class. Class-balanced
thresholding strategy \citep{zou2018unsupervised} selects a certain
percentile ($q_{M}$) of the prediction confidences. We choose a low
percentile value to obtain highly confident predictions. Threshold
values are calculated in training time with the moving average of
the percentile of the prediction of the current frame.

One successful approach in proxy-based metric learning is to use neighborhood
component analysis (NCA) in training, where the samples are compared
against proxies. The proxy features and the metric network parameters
are updated concurrently to attract the feature of a sample to the
corresponding proxy feature and repel from the other proxy features.
We apply a similar approach to train the metric network and proxy
features. Teh et al.
\citep{teh2020proxynca++} show that using small temperature $T$ values refine
decision boundaries and help classify samples better. Motivated by
these, we utilize NCA with temperature scaling. We select an equal
number of samples in each class in $\tilde{y}_{M}^{h,w,c}$ to ensure
balanced training. The metric network is optimized the following loss
function:

\begin{equation}
\mathcal{L}_{M}=\sum_{h,w}\left[-\log\left(\frac{\exp(-d(f^{h,w},f_{p})*\frac{1}{T})}{\sum_{c}\exp(-d(f^{h,w},f_{p_{c}})*\frac{1}{T})}\right)\right],
\end{equation}

\noindent where $T$ is a temperature and $d(x,y)$ is a normalized
squared L2-Norm:

\begin{equation}
d(x,y)=\sum_{i}\left(\frac{x_{i}}{\left\lVert x\right\rVert }-\frac{y_{i}}{\left\lVert y\right\rVert }\right)^{2}
\end{equation}

The same distance function is used in both training the metric network
and calculating the reliability score. The smaller the distance gets, the more similar to
the corresponding class it becomes. In order to calculate the similarity,
we first feed the features of the feature extractor network of the
teacher model to the metric network. The metric network outputs a
metric feature for each pixel of the input image of the teacher network.
Then we generate the similarity map by calculating the distance between
the metric features and the proxy feature of the class predicted by
the teacher network. We use a reverse sigmoid function to transform
similarity to the reliability, where $\alpha$ is sharpness constant,
and $\beta$ is offset of the sigmoid function: 

\begin{equation}
w^{(h,w)}=\frac{1}{1+e^{-\alpha*(\beta-d(f^{h,w},\hat{f_{p}}))}}
\end{equation}

\subsection{Metric-based online ClassMix}

The noise injection to the input of the student model leads to a better
generalization for both the student and teacher model \citep{xie2020self}.
Data augmentation with photometric noise is a common approach in computer
vision applications. Another effective augmentation technique for
classification and semantic segmentation is the mixing method. The
mixing methods combine pixels from two training images to create a
highly perturbed sample. ClassMix algorithm \citep{olsson2021classmix}
is a mixing data augmentation method that cuts half of the classes
in the predicted image and pastes on the other image. 

In the context of source-free domain adaptation, the unavailability of the source dataset restricts the ability to mix target data with the source data. Additionally, the absence of labels in the target dataset prevents the extraction of class patches from the target images based on groundtruth information. To address this challenge, we propose a method called Metric-based Online ClassMix (MOCM). This method effectively stores reliable patches during the training phase using metric distance and subsequently incorporates these patches into the input data.

We maintain separate patch buffers for each class. These buffers have a fixed size and operate on a first-in-first-out scheme, effectively reducing memory requirements while ensuring that the most up-to-date patches are stored. To select which patches to store, we utilize the distance between the metric feature predicted by the metric network and the proxy features. If the average distance between the metric features and the corresponding class proxy ($f_{p}^{c}$) of a patch falls below a predefined threshold ($\tau_{MOCM}$), the patch is added to the buffer of class $c$. Since the mixing operation occurs in both image and label space, we store the image, pseudo-label, and reliability map patch tuple in the buffer.

To utilize the stored patches, we first apply a photometric noise
to the input image, then apply the proposed MOCM method. We sample
one patch from each $N_{MOCM}$ classes randomly and paste them onto
the image ($x_{CM}$), pseudo-label ($\tilde{y}_{CM}$), and reliability
map ($w_{MOCM}$). The training of the student network is performed
by the stochastic gradient descent in order to minimize the following
loss function:

\begin{equation}
\begin{aligned}\mathcal{L}=\frac{1}{H\times W}\sum\Biggl[\Biggl( & -\sum_{c}\tilde{y}_{MOCM}^{h,w,c}\times\log(T(x_{MOCM})^{h,w,c})\Biggr)\\
 & \circ w_{MOCM}^{h,w}\Biggr].
\end{aligned}
\label{eq:final-loss}
\end{equation}

\section{Experiments}

\begin{table*}
\centering

\caption{Comparison with state-of-the-art on GTAV-to-CityScapes in terms of
per-class IoUs and mIoU (\%). SF represents if the method is in the
source-free setting.}

\scalebox{0.65}{%
\centering
\begin{tabular}{l|c|c|ccccccccccccccccccc|c}
Method & \rotatebox{90}{Network} & SF & \rotatebox{90}{Road} & \rotatebox{90}{Sidewalk} & \rotatebox{90}{Building} & \rotatebox{90}{Wall} & \rotatebox{90}{Fence} & \rotatebox{90}{Pole} & \rotatebox{90}{Light} & \rotatebox{90}{Sign} & \rotatebox{90}{Veg.} & \rotatebox{90}{Terrain} & \rotatebox{90}{Sky} & \rotatebox{90}{Person} & \rotatebox{90}{Rider} & \rotatebox{90}{Car} & \rotatebox{90}{Truck} & \rotatebox{90}{Bus} & \rotatebox{90}{Train} & \rotatebox{90}{Mbike} & \rotatebox{90}{Bicycle} & \rotatebox{90}{mIoU}\tabularnewline
\hline 
AdvEnt \citep{vu2019advent} & \multirow{13}{*}{\rotatebox{90}{DeepLabV2}} & \xmark & 89.4 & 33.1 & 81.0 & 26.6 & 26.8 & 27.2 & 33.5 & 24.7 & 83.9 & 36.7 & 78.8 & 58.7 & 30.5 & 84.8 & 38.5 & 44.5 & 1.7 & 31.6 & 32.4 & 45.5\tabularnewline
Intra-domain \citep{pan2020unsupervised} &  & \xmark & 90.6 & 37.1 & 82.6 & 30.1 & 19.1 & 29.5 & 32.4 & 20.6 & 85.7 & 40.5 & 79.7 & 58.7 & 31.1 & 86.3 & 31.5 & 48.3 & 0.0 & 30.2 & 35.8 & 45.8\tabularnewline
MaxSquare \citep{chen2019domain} &  & \xmark & 89.4 & 43.0 & 82.1 & 30.5 & 21.3 & 30.3 & 34.7 & 24.0 & 85.3 & 39.4 & 78.2 & \uline{63.0} & 22.9 & 84.6 & 36.4 & 43.0 & 5.5 & 34.7 & 33.5 & 46.4\tabularnewline
LSE + FL \citep{subhani2020learning} &  & \xmark & 90.2 & 40.0 & 83.5 & 31.9 & 26.4 & 32.6 & 38.7 & 37.5 & 81.0 & 34.2 & 84.6 & 61.6 & \uline{33.4} & 82.5 & 32.8 & 45.9 & 6.7 & 29.1 & 30.6 & 47.5\tabularnewline
BDL \citep{li2019bidirectional} &  & \xmark & 91.0 & 44.7 & 84.2 & 34.6 & 27.6 & 30.2 & 36.0 & 36.0 & 85.0 & 43.6 & 83.0 & 58.6 & 31.6 & 83.3 & 35.3 & 49.7 & 3.3 & 28.8 & 35.6 & 48.5\tabularnewline
Stuff and Things \citep{wang2020differential} &  & \xmark & 90.6 & 44.7 & 84.8 & 34.3 & 28.7 & 31.6 & 35.0 & 37.6 & 84.7 & 43.3 & 85.3 & 57.0 & 31.5 & 83.8 & 42.6 & 48.5 & 1.9 & 30.4 & 39.0 & 49.2\tabularnewline
Texture Invariant \citep{kim2020learning} &  & \xmark & \textbf{92.9} & 55.0 & 85.3 & 34.2 & 31.1 & 34.9 & 40.7 & 34.0 & 85.2 & 40.1 & 87.1 & 61.0 & 31.1 & 82.5 & 32.3 & 42.9 & 0.3 & 36.4 & 46.1 & 50.2\tabularnewline
FDA \citep{yang2020fda} &  & \xmark & 92.5 & 53.3 & 82.4 & 26.5 & 27.6 & \uline{36.4} & 40.6 & 38.9 & 82.3 & 39.8 & 78.0 & 62.6 & \textbf{34.4} & 84.9 & 34.1 & 53.1 & \textbf{16.9} & 27.7 & 46.4 & 50.4\tabularnewline
DMLC \citep{guo2021metacorrection} &  & \xmark & \uline{92.8} & \textbf{58.1} & 86.2 & 39.7 & \uline{33.1} & 36.3 & \textbf{42.0} & 38.6 & 85.5 & 37.8 & \textbf{87.6} & 62.8 & 31.7 & 84.8 & 35.7 & 50.3 & 2.0 & 36.8 & 48.0 & 52.1\tabularnewline
URMA \citep{fleuret2021uncertainty} &  & \cmark & 92.3 & \uline{55.2} & 81.6 & 30.8 & 18.8 & \textbf{37.1} & 17.7 & 12.1 & 84.2 & 35.9 & 83.8 & 57.7 & 24.1 & 81.7 & 27.5 & 44.3 & \uline{6.9} & 24.1 & 40.4 & 45.1\tabularnewline
LD \citep{you2021domain} &  & \cmark & 91.6 & 53.2 & 80.6 & 36.6 & 14.2 & 26.4 & 31.6 & 22.7 & 83.1 & 42.1 & 79.3 & 57.3 & 26.6 & 82.1 & 41.0 & 50.1 & 0.3 & 25.9 & 19.5 & 45.5\tabularnewline
DT+AC \citep{yang2022source} & & \cmark & 78.0 & 29.5 & 83.0 & 29.3 & 21.0 & 31.8 & 38.1 & 33.1 & 83.8 & 39.2 & 80.8 & 61.0 & 30.0 & 83.9 & 26.1 & 40.4 & 1.9 & 34.2 & 43.7 & 45.7\tabularnewline
SRDA \citep{bateson2020source} &  & \cmark & 90.5 & 47.1 & 82.8 & 32.8 & 28.0 & 29.9 & 35.9 & 34.8 & 83.3 & 39.7 & 76.1 & 57.3 & 23.6 & 79.5 & 30.7 & 40.2 & 0.0 & 26.6 & 30.9 & 45.8\tabularnewline
HCL \citep{huang2021model} & & \cmark & 92.0 & 55.0 & 80.4 & 33.5 & 24.6 & \textbf{37.1} & 35.1 & 28.8 & 83.0 & 37.6 & 82.3 & 59.4 & 27.6 & 83.6 & 32.3 & 36.6 & 14.1 & 28.7 & 43.0 & 48.1\tabularnewline
\cline{1-1} \cline{3-23} \cline{4-23} \cline{5-23} \cline{6-23} \cline{7-23} \cline{8-23} \cline{9-23} \cline{10-23} \cline{11-23} \cline{12-23} \cline{13-23} \cline{14-23} \cline{15-23} \cline{16-23} \cline{17-23} \cline{18-23} \cline{19-23} \cline{20-23} \cline{21-23} \cline{22-23} \cline{23-23} 
STvM (w/o MST) &  & \cmark & 91.4 & 52.9 & \uline{87.3} & \uline{41.5} & \textbf{33.3} & 35.9 & 40.8 & \uline{48.5} & \uline{87.3} & \uline{49.2} & \uline{87.4} & 62.2 & 9.3 & \textbf{87.1} & \uline{45.5} & \uline{58.7} & 0.0 & \uline{47.4} & \uline{59.8} & \uline{54.0}\tabularnewline
STvM (w/ MST) &  & \cmark & 92.1 & 55.1 & \textbf{87.6} & \textbf{45.5} & \textbf{33.3} & 36.1 & \uline{41.8} & \textbf{48.6} & \textbf{87.9} & \textbf{51.1} & \textbf{87.6} & \textbf{63.2} & 8.6 & \uline{87.0} & \textbf{47.8} & \textbf{59.8} & 0.0 & \textbf{50.1} & \textbf{60.4} & \textbf{54.9}\tabularnewline
\hline 
MinEnt \citep{vu2019advent} & \multirow{7}{*}{\rotatebox{90}{DeepLabV3}} & \xmark & 80.2 & 31.9 & 81.4 & 25.1 & 20.8 & 24.6 & 30.2 & 17.5 & 83.2 & 18.0 & 76.2 & 55.2 & 24.6 & 75.5 & 33.2 & 31.2 & 4.4 & 27.4 & 22.9 & 40.2\tabularnewline
AdaptSegNet \citep{tsai2018learning} &  & \xmark & 81.6 & 26.6 & 79.5 & 20.7 & 20.5 & 23.7 & 29.9 & 22.6 & 81.6 & 26.7 & 81.2 & 52.4 & 20.2 & 79.1 & \textbf{36.0} & 28.8 & \textbf{7.5} & 24.7 & 26.2 & 40.5\tabularnewline
CBST \citep{zou2018unsupervised} &  & \xmark & 84.8 & 41.5 & 80.4 & 19.5 & 22.4 & 24.7 & 30.2 & 20.4 & 83.5 & 29.6 & 82.3 & 54.7 & 25.3 & 79.2 & 34.5 & 32.3 & \uline{6.8} & \uline{29.0} & \textbf{34.9} & 42.9\tabularnewline
MaxSquare \citep{chen2019domain} &  & \xmark & 85.8 & 33.6 & 82.4 & 25.3 & 25.0 & 26.5 & 33.3 & 18.7 & 83.2 & 32.9 & 79.8 & 57.8 & 22.2 & 81.0 & 32.1 & 32.6 & 5.2 & \textbf{29.8} & \uline{32.4} & 43.1\tabularnewline
SFDA \citep{liu2021source} &  & \cmark & 84.2 & 39.2 & 82.7 & 27.5 & 22.1 & 25.9 & 31.1 & 21.9 & 82.4 & 30.5 & 85.3 & 58.7 & 22.1 & 80.0 & 33.1 & 31.5 & 3.6 & 27.8 & 30.6 & 43.2\tabularnewline
\cline{1-1} \cline{3-23} \cline{4-23} \cline{5-23} \cline{6-23} \cline{7-23} \cline{8-23} \cline{9-23} \cline{10-23} \cline{11-23} \cline{12-23} \cline{13-23} \cline{14-23} \cline{15-23} \cline{16-23} \cline{17-23} \cline{18-23} \cline{19-23} \cline{20-23} \cline{21-23} \cline{22-23} \cline{23-23} 
STvM (w/o MST) &  & \cmark & \uline{90.3} & \uline{50.2} & \uline{87.4} & \uline{37.9} & \textbf{33.0} & \uline{35.8} & \uline{45.2} & \uline{48.5} & \uline{85.7} & \textbf{44.1} & \uline{86.1} & \uline{62.4} & \uline{29.8} & \uline{84.3} & 30.2 & \uline{50.0} & 0.6 & 7.4 & 0.0 & \uline{47.8}\tabularnewline
STvM (w/ MST) &  & \cmark & \textbf{90.8} & \textbf{50.7} & \textbf{87.7} & \textbf{40.9} & \uline{32.0} & \textbf{36.1} & \textbf{47.2} & \textbf{48.6} & \textbf{85.9} & \uline{43.3} & \textbf{87.0} & \textbf{62.9} & \textbf{31.5} & \textbf{85.2} & \uline{34.6} & \textbf{54.0} & 0.6 & 7.8 & 0.0 & \textbf{48.8}\tabularnewline
\hline 
\end{tabular}}
\label{tab:gta_results}
\end{table*}

\subsection{Datasets}

We demonstrate the performance of STvM on three different source-to-target
adaptation scenarios which are GTA5-to-Cityscapes, Synthia-to-Cityscapes,
and Cityscapes-to-NTHU Cross-City.

The Cityscapes \citep{cordts2016cityscapes} dataset consists of 5000
real-world street-view images captured in 50 different cities. The
images have high-quality pixel-level annotations with a resolution
of 2048 x 1024. They are annotated with 19 semantic labels for semantic
segmentation. The Cityscapes is split into training, validation, and
test sets containing 2975, 500, and 1525 images, respectively. The
methods are evaluated on the validation set following the standard
setting used in the previous domain adaptation studies. The GTA5 \citep{Richter_2016_ECCV}
is a synthetic dataset where images and labels are automatically grabbed
from Grand Theft Auto V video game. It consists of 24,966 synthetic
images with the size of 1914 x 1052. They have pixel-level annotations
of 33 categories. The 19 classes compatible with the Cityscapes are
used in the experiments. The Synthia\footnote{This dataset is subject to the CC-BY-NC-SA 3.0}
\citep{Ros_2016_CVPR} is also a synthetic dataset. It is composed
of urban scene images with pixel-level annotations. The commonly used
SYNTHIA-RAND-CITYSCAPES subset contains 9,400 images with a resolution
of 1280 x 760. The dataset has 16 shared categories with the Cityscapes
dataset. Cross-City \citep{chen2017no} is a real-world dataset. The
images are recorded in four cities, which are Rome, Rio, Taipei, and
Tokyo. The dataset has 3200 unlabeled images as a training set and
100 labeled images for the test set with the size of 2048 x 1024.
Cross-City has 13 shared classes with the Cityscapes dataset.

\subsection{Implementation details}

\begin{table*}
\centering

\caption{Comparison with state-of-the-art on SYNTHIA-to-CityScapes in terms
of per-class IoUs and mIoU (\%). The mIoU{*} column denotes the mean
IoU over 13 categories excluding those marked by {*}. SF represents
if the method is in the source-free setting.}

\scalebox{0.65}{\setlength{\tabcolsep}{7.55pt}%
\begin{tabular}{l|c|c|cccccccccccccccc|c|c}
Method & \rotatebox{90}{Network} & SF & \rotatebox{90}{Road} & \rotatebox{90}{Sidewalk} & \rotatebox{90}{Building} & \rotatebox{90}{Wall{*}} & \rotatebox{90}{Fence{*}} & \rotatebox{90}{Pole{*}} & \rotatebox{90}{Light} & \rotatebox{90}{Sign} & \rotatebox{90}{Veg.} & \rotatebox{90}{Sky} & \rotatebox{90}{Person} & \rotatebox{90}{Rider} & \rotatebox{90}{Car} & \rotatebox{90}{Bus} & \rotatebox{90}{Mbike} & \rotatebox{90}{Bicycle} & \rotatebox{90}{mIoU} & \rotatebox{90}{mIoU{*}}\tabularnewline
\hline 
AdvEnt \citep{vu2019advent} & \multirow{12}{*}{\rotatebox{90}{DeepLabV2}} & \xmark & 85.6 & 42.2 & 79.7 & 8.7 & 0.4 & 25.9 & 5.4 & 8.1 & 80.4 & 84.1 & 57.9 & 23.8 & 73.3 & 36.4 & 14.2 & 33.0 & 41.2 & 48.0\tabularnewline
MaxSquare \citep{chen2019domain} &  & \xmark & 82.9 & 40.7 & 80.3 & \uline{10.2} & 0.8 & 25.8 & 12.8 & 18.2 & 82.5 & 82.2 & 53.1 & 18.0 & 79.0 & 31.4 & 10.4 & 35.6 & 41.5 & 48.2\tabularnewline
Intra-domain \citep{pan2020unsupervised} &  & \xmark & 84.3 & 37.7 & 79.5 & 5.3 & 0.4 & 24.9 & 9.2 & 8.4 & 80.0 & 84.1 & 57.2 & 23.0 & 78.0 & 38.1 & 20.3 & 36.5 & 41.7 & 48.9\tabularnewline
Texture Invariant \citep{kim2020learning} &  & \xmark & \textbf{92.6} & \textbf{53.2} & 79.2 & - & - & - & 1.6 & 7.5 & 78.6 & 84.4 & 52.6 & 20.0 & 82.1 & 34.8 & 14.6 & 39.4 & - & 49.3\tabularnewline
LSE + FL \citep{subhani2020learning} &  & \xmark & 82.9 & 43.1 & 78.1 & 9.3 & 0.6 & 28.2 & 9.1 & 14.4 & 77.0 & 83.5 & 58.1 & 25.9 & 71.9 & 38.0 & \uline{29.4} & 31.2 & 42.5 & 49.4\tabularnewline
BDL \citep{li2019bidirectional} &  & \xmark & \uline{86.0} & 46.7 & 80.3 & - & - & - & 14.1 & 11.6 & 79.2 & 81.3 & 54.1 & 27.9 & 73.7 & 42.2 & 25.7 & 45.3 & - & 51.4\tabularnewline
Stuff and Things \citep{wang2020differential} &  & \xmark & 83.0 & 44.0 & 80.3 & - & - & - & 17.1 & 15.8 & 80.5 & 81.8 & 59.9 & \textbf{33.1} & 70.2 & 37.3 & 28.5 & 45.8 & - & 52.1\tabularnewline
FDA \citep{yang2020fda} &  & \xmark & 79.3 & 35.0 & 73.2 & - & - & - & 19.9 & 24.0 & 61.7 & 82.6 & \textbf{61.4} & \uline{31.1} & 83.9 & 40.8 & \textbf{38.4} & 51.1 & - & 52.5\tabularnewline
DMLC \citep{guo2021metacorrection} &  & \xmark & \textbf{92.6} & \uline{52.7} & 81.3 & 8.9 & \textbf{2.4} & 28.1 & 13.0 & 7.3 & 83.5 & 85.0 & \uline{60.1} & 19.7 & 84.8 & 37.2 & 21.5 & 43.9 & 45.1 & 52.5\tabularnewline
URMA \citep{fleuret2021uncertainty} &  & \cmark & 59.3 & 24.6 & 77.0 & \textbf{14.0} & \uline{1.8} & 31.5 & 18.3 & 32.0 & 83.1 & 80.4 & 46.3 & 17.8 & 76.7 & 17.0 & 18.5 & 34.6 & 39.6 & 45.0\tabularnewline
DT+AC \citep{yang2022source} & & \cmark & 77.5 & 37.4 & 80.5 & 13.5 & 1.7 & 30.5 & 24.8 & 19.7 & 79.1 & 83.0 & 49.1 & 20.8 & 76.2 & 12.1 & 16.5 & 46.1 & 41.8 & 47.9\tabularnewline
LD \citep{you2021domain} & & \cmark & 77.1 & 33.4 & 79.4 & 5.8 & 0.5 & 23.7 & 5.2 & 13.0 & 81.8 & 78.3 & 56.1 & 21.6 & 80.3 & \textbf{49.6} & 28.0 & 48.1 & 42.6 & 50.1\tabularnewline
HCL \citep{huang2021model} & & \cmark & 80.9 & 34.9 & 76.7 & 6.6 & 0.2 & \uline{36.1} & 20.1 & 28.2 & 79.1 & 83.1 & 55.6 & 25.6 & 78.8 & 32.7 & 24.1 & 32.7 & 43.5 & 50.2\tabularnewline
\cline{1-1} \cline{3-21} \cline{4-21} \cline{5-21} \cline{6-21} \cline{7-21} \cline{8-21} \cline{9-21} \cline{10-21} \cline{11-21} \cline{12-21} \cline{13-21} \cline{14-21} \cline{15-21} \cline{16-21} \cline{17-21} \cline{18-21} \cline{19-21} \cline{20-21} \cline{21-21} 
STvM (w/o MST) &  & \cmark & 71.7 & 31.0 & \uline{83.7} & 0.2 & 0.1 & 35.8 & \uline{34.7} & \uline{37.9} & \uline{84.8} & \uline{87.5} & 53.6 & 23.0 & \uline{85.8} & 46.7 & 26.9 & \uline{52.6} & \uline{47.2} & \uline{55.4}\tabularnewline
STvM (w/ MST) &  & \cmark & 71.6 & 31.1 & \textbf{84.2} & 0.0 & 0.0 & \textbf{36.7} & \textbf{36.0} & \textbf{38.4} & \textbf{85.4} & \textbf{87.8} & 55.4 & 23.5 & \textbf{85.9} & \textbf{47.8} & 29.1 & \textbf{53.6} & \textbf{47.9} & \textbf{56.1}\tabularnewline
\hline 
MinEnt\citep{vu2019advent} & \multirow{7}{*}{\rotatebox{90}{DeepLabV3}} & \xmark & 78.2 & 39.6 & 81.9 & 4.3 & 0.2 & 26.2 & 2.2 & 4.1 & 81.1 & 87.7 & 37.7 & 7.2 & 75.8 & 24.9 & 4.6 & 25.1 & 36.3 & 42.3\tabularnewline
AdaptSegNet \citep{tsai2018learning} &  & \xmark & 79.7 & 38.6 & 79.3 & 5.6 & 0.8 & 25.4 & 3.6 & 5.5 & 80.0 & 85.4 & 40.8 & 11.7 & 79.8 & 21.4 & 5.2 & 30.5 & 37.1 & 43.2\tabularnewline
CBST \citep{zou2018unsupervised} &  & \xmark & \uline{81.4} & \uline{44.2} & 80.4 & 7.9 & 0.7 & 25.6 & 5.2 & 12.4 & 81.4 & \uline{89.5} & 39.7 & 10.6 & \uline{82.1} & 21.9 & 6.3 & 32.9 & 38.9 & 45.2\tabularnewline
MaxSquare \citep{chen2019domain} &  & \xmark & 81.0 & 39.8 & 82.6 & 8.7 & 0.5 & 23.2 & 6.6 & 12.4 & \uline{85.3} & \textbf{90.1} & 39.9 & 8.4 & \textbf{84.7} & 19.4 & 10.2 & 33.4 & 39.1 & 45.7\tabularnewline
SFDA \citep{liu2021source} &  & \cmark & \textbf{81.9} & \textbf{44.9} & 81.7 & 4.0 & 0.5 & 26.2 & 3.3 & 10.7 & \textbf{86.3} & 89.4 & 37.9 & 13.4 & 80.6 & 25.6 & 9.6 & 31.3 & 39.2 & 45.9\tabularnewline
\cline{1-1} \cline{3-21} \cline{4-21} \cline{5-21} \cline{6-21} \cline{7-21} \cline{8-21} \cline{9-21} \cline{10-21} \cline{11-21} \cline{12-21} \cline{13-21} \cline{14-21} \cline{15-21} \cline{16-21} \cline{17-21} \cline{18-21} \cline{19-21} \cline{20-21} \cline{21-21} 
STvM (w/o MST) &  & \cmark & 49.3 & 20.7 & \uline{84.2} & \uline{14.0} & \uline{1.2} & \uline{33.1} & \uline{42.4} & \uline{45.9} & 82.9 & 82.8 & \uline{50.4} & \uline{22.1} & 81.1 & \uline{38.4} & \uline{21.2} & \uline{46.5} & \uline{44.8} & \uline{51.4}\tabularnewline
STvM (w/ MST) &  & \cmark & 45.9 & 19.8 & \textbf{84.5} & \textbf{15.3} & \textbf{1.3} & \textbf{33.7} & \textbf{44.6} & \textbf{46.5} & 83.3 & 84.0 & \textbf{51.3} & \textbf{23.1} & 80.9 & \textbf{40.7} & \textbf{24.8} & \textbf{47.2} & \textbf{45.4} & \textbf{52.1}\tabularnewline
\hline 
\end{tabular}}\label{tab:synthia_results}
\end{table*}

To ensure a fair comparison with previous unsupervised source-free domain adaptation methods for semantic segmentation, we employ two different segmentation networks. The first network is DeepLabV2 \citep{chen2017deeplab} with the ResNet-101 backbone, and the second network is DeepLabV3 \citep{chen2017rethinking} with the ResNet-50 backbone. Both the teacher ($\mathcal{T}$) and the student ($\mathcal{S}$) networks adopt the same architecture. ResNet-101 and ResNet-50 are utilized as feature extractors, while the ASPP module of the DeepLabV2 or DeepLabV3 architecture is employed for the classifier ($\mathcal{C}$) and metric ($\mathcal{M}$) networks, depending on the chosen architecture.

We perform the experiments on a single NVIDIA RTX 2080 Ti GPU using
the PyTorch framework \citep{paszke2017automatic}. We use the same
parameters in the training of both DeepLabV2 and DeepLabV3. We resize
the input image to $1024\times512$ and cropped $512\times512$ patch
randomly in our experiments. The batch size is set to $2$.

Both the student and the teacher network is initialized with the source
model which is trained in the source domain. The student network ($\mathcal{S}$)
is trained with the Stochastic Gradient Descent (SGD) \citep{bottou2010large}
optimizer with Nesterov acceleration. The initial learning rate is
set to $2.5\times10^{-4}$ and $2.5\times10^{-3}$ for the feature
extractor ($\mathcal{F}$) and the classifier ($\mathcal{C}$), respectively.
The momentum is set to $0.9$, and the weight decay is set to $5.0\times10^{-4}$.
The teacher network is updated once in $100$ iterations with the
parameters of the student network, setting the smoothing factor as
$1.0\times10^{-3}$. The Adam \citep{DBLP:journals/corr/KingmaB14}
optimizer is utilized for training the metric network ($\mathcal{M}$)
with the initial learning rate of $3.0\times10^{-4}$. The learning
rates of both optimizers are scheduled with polynomial weight decay
with the power of $0.9$. All the networks are trained concurrently,
enabling end-to-end training.

As for the hyper-parameters, metric feature size $N_{f}$, metric
temperature $T$, metric pseudo-label threshold percentile $q_{m}$
and the patch buffer size are set to $128$, $0.25$, $0.2$ and $50$,
respectively. The metric distance to reliability transformation parameters
$\alpha$ and $\beta$ are set to $2$ and $0.6$. The metric distance
threshold $\tau_{MOCM}$ is set to $0.8$.

\subsection{Results}

\begin{table}
\centering\caption{Comparison with state-of-the-art on CityScapes-to-NTHU in terms of
mIoU (\%). DL V2 and DL V3 represents DeepLabV2 with ResNet-101 backbone
and DeepLabV3 with ResNet-50 backbone, respectively.}

\scalebox{0.85}{%
\begin{tabular}{l|c|c|c|c|c|c}
Method & \rotatebox{90}{Network} & \rotatebox{90}{Rome} & \rotatebox{90}{Rio} & \rotatebox{90}{Tokyo} & \rotatebox{90}{Taipei} & \rotatebox{90}{Mean}\tabularnewline
\hline 
URMA \citep{fleuret2021uncertainty} & \multirow{3}{*}{\rotatebox{90}{DL V2}} & \uline{53.8} & 53.5 & 49.8 & 50.1 & 51.8\tabularnewline
STvM (w/o MST) &  & 53.1 & \uline{54.5} & \uline{51.6} & \uline{51.4} & \uline{52.6}\tabularnewline
STvM (w/ MST) &  & \textbf{54.2} & \textbf{56.8} & \textbf{52.7} & \textbf{53.4} & \textbf{54.3}\tabularnewline
\hline 
SFDA \citep{liu2021source} & \multirow{3}{*}{\rotatebox{90}{DL V3}} & 48.3 & 49.0 & \textbf{46.4} & \textbf{47.2} & 47.7\tabularnewline
STvM (w/o MST) &  & \uline{50.8} & \uline{51.0} & \uline{46.1} & 45.0 & \uline{48.2}\tabularnewline
STvM (w/ MST) &  & \textbf{51.2} & \textbf{52.6} & \textbf{46.4} & \uline{46.0} & \textbf{49.0}\tabularnewline
\hline 
\end{tabular}}

\label{tab:nthu_results}
\end{table}

We evaluated our proposed method STvM on two challenging synthetic-to-real
and one cross-city domain adaptation scenarios, namely GTA5-to-CityScapes,
SYNTHIA-to-CityScapes, and CityScapes-to-Cross-City settings. We compared
our method with the state-of-the-art methods, containing both classical
and source-free methods. The results, obtained with two network architectures,
are given in Table \ref{tab:gta_results}, \ref{tab:synthia_results},
and \ref{tab:nthu_results}. MST corresponds to multi-scale testing. 

Classical domain adaptation methods, utilizing the labeled source-domain
dataset alongside the unlabeled target dataset, gain the advantage
of better convergence than source-free domain adaptation methods.
Therefore, they usually perform better than source-free settings. 

STvM is a source-free domain adaptation method. The experimental results
show that our method outperforms the state-of-the-art methods in the
same class by a large margin. Specifically, It improves the performance
of SRDA by 20\% on the GTA5-to-CityScapes dataset with DeepLabV2,
SFDA by \%14 on SYNTHIA-to-CityScapes dataset with DeepLabV3, and
SFDA by 13\% on the GTA5-to-CityScapes dataset with DeepLabV3. We
observed that metric learning is a powerful tool for discriminating
confusing classes such as building/wall and light/sign. In addition,
STvM shows better or comparable performance with the classical domain
adaptation methods.

\subsubsection{Ablation study}

In order to analyze the effect of each component on the performance,
we present the results of the ablation study in Table \ref{tab:ablation_study}.
We named the methods as $Source$, $ST$, $ST_{Aug}$, $ST_{MT}$,
$STvM_{Raw}$, and $STvM$. $Source$ is a model trained in source-domain
without any domain adaptation method is applied. $ST$ model represents
a network trained with the self-training (ST) method using all pseudo-labels.
Conforming the common knowledge, the self-training boosts the performance
of the source model by +6.8\%. $ST_{Aug}$utilizes both self-training
(ST) and photometric augmentation (Aug.) to the input image. Injection
of the photometric noise to the self-training provides +2.1\% improvement.
$ST_{MT}$ follows the mean teacher (MT) approach. All predictions
of the teacher network are used as pseudo-labels to train the student
network. While photometric augmentation is applied to the input of
the student, no augmentation is applied to the input of the teacher
network. Enabling mean-teacher with self-training by using all the
predictions of the teacher model contributes +2.4\%. As an extension
to the $ST_{MT}$, $STvM_{Raw}$ benefits from the gradient scaling (GS). The metric network is used to estimate the reliability of the predictions of the
teacher network. The metric network that
generates reliability to scale gradients of the predictions improves
performance by +1.4\% showing the effectiveness of the proposed method.
$STvM$ also exploits the metric-based online ClassMix (MOCM). Using
the metric distance in the generation of the patches improves performance
significantly. Generating patches and mixing with the input in training
time using the reliability has a strong positive impact on the performance
by +4.7\%.

\begin{table}[h]
\centering\caption{Ablation Study on GTAV-to-CityScapes}

\scalebox{1.0}{%
\begin{tabular}{l|c|c|c|c|c|c|c}
Name & \rotatebox{90}{ST} & \rotatebox{90}{Aug.} & \rotatebox{90}{MT} & \rotatebox{90}{MGS} & \rotatebox{90}{MOCM} & mIoU & $\Delta$\tabularnewline
\hline 
$Source$ & - & - & - & - & - & 36.6 & -\tabularnewline
$ST$ & \cmark & - & - & - & - & 43.4 & +6.8\tabularnewline
$ST_{Aug}$ & \cmark & \cmark & - & - & - & 45.5 & +2.1\tabularnewline
$ST_{MT}$ & \cmark & \cmark & \cmark & - & - & 47.9 & +2.4\tabularnewline
$STvM_{Raw}$ & \cmark & \cmark & \cmark & \cmark & - & 49.3 & +1.4\tabularnewline
$STvM$ & \cmark & \cmark & \cmark & \cmark & \cmark & \textbf{54.0} & \textbf{+4.7}\tabularnewline
\hline 
\end{tabular}}

\label{tab:ablation_study}
\end{table}

\subsubsection{Hyper-Parameter analysis}

In order to evaluate the influence of the hyper-parameters, we conduct
experiments for four different values of all hyper-parameters: MOCM
threshold ($\tau_{MOCM}$), MOCM patch per image ($N_{MOCM}$), metric
feature size ($N_{f}$), metric proxy temperature ($T$), metric pseudo-label
quantile ($q_{M}$), reverse sigmoid alpha ($\alpha$), reverse sigmoid
beta ($\beta$). The experimental results are given in Table \ref{tab:hyperparameter_analysis}.
The parameter stated in the name column is modified in the experiments.
The default values are kept for the rest. We observed that no parameter
has a noteworthy impact on the performance except large $\alpha$
and small $N_{MOCM}$ values. Assigning large $\alpha$ leads to a
sharp decrease in the reliability of the sample that limits the contribution
of the samples that are far away to the corresponding proxy. This
is a desirable situation for the perfectly trained metric network.
However, the metric network is trained with highly-confident predictions
of the teacher network. Therefore, the metric network is overconfident
for the easy samples. If large $\alpha$ is selected, the student
network is biased towards easy samples. That decreases diversity and
has a negative impact on performance. Using a small $N_{MOCM}$ value
decreases the complexity of the augmentation, limiting the contribution
of the proposed metric-based online ClassMix method.

\begin{table}[h]
\centering

\caption{HyperParameter Analysis on GTAV-to-CityScapes}

\scalebox{1.0}{%
\begin{tabular}{l|c|c|c|c}
Name & \multicolumn{4}{c}{Parameter Value / Performance}\tabularnewline
\hline 
$\tau_{MOCM}$ & 0.4 / 52.9 & 0.6 / 53.07 & 0.8 / 54.0 & 1.0 / 53.0\tabularnewline
$N_{MOCM}$ & 2 / 51.1 & 5 / 52.4 & 10 / 54.0 & 15 / 53.5\tabularnewline
$N_{f}$ & 32 / 53.1 & 64 / 53.0 & 128 / 54.0 & 256 / 52.9\tabularnewline
$T$ & 0.1 / 53.8 & 0.25 / 54.0 & 0.5 / 52.9 & 1.0 / 52.3\tabularnewline
$q_{M}$ & 0.1 / 52.6 & 0.2 / 54.0 & 0.3 / 53.5 & 0.4 / 53.3\tabularnewline
$\alpha$ & 1 / 53.1 & 2 / 54.0 & 4 / 51.3 & 8 / 50.0\tabularnewline
$\beta$ & 0.5 / 53.0 & 0.6 / 54.0 & 0.7 / 53.6 & 0.8 / 53.8\tabularnewline
\hline 
\end{tabular}}

\label{tab:hyperparameter_analysis}
\end{table}

\subsubsection{Variance analysis}

We made a variance analysis for our STvM and compared it to state-of-the-art
methods. Following URMA \citep{fleuret2021uncertainty}, we conduct
five experiments with different random seeds, keeping the hyper-parameters
in their default values. The mean and standard deviations obtained
for GTA5-to-CityScapes with DeepLabV2 are reported in Table \ref{tab:variance_analysis}.
Experiments show that STvM shows robust performance with the lowest
standard deviation among the state-of-the-art methods.

\begin{table}[h]
\centering

\caption{Variance Analysis on GTAV-to-CityScapes}

\scalebox{1.0}{%
\begin{tabular}{l|c|c}
Method & Performance Estimate & Min\tabularnewline
\hline 
AdaptSegnet & 39.68 \textpm{} 1.49 & 37.70\tabularnewline
ADVENT & 42.56 \textpm{} 0.64 & 41.60\tabularnewline
CBST & 44.04 \textpm{} 0.88 & 42.80\tabularnewline
UMRA & 42.44 \textpm{} 2.18 & 39.71\tabularnewline
STvM & \textbf{53.55 \textpm{} 0.25} & \textbf{53.26}\tabularnewline
\hline 
\end{tabular}}

\label{tab:variance_analysis}
\end{table}

\subsubsection{Metric network evaluation}

The metric network clusters classes in the metric feature space, providing
a reliable distance measure. The silhouette score is a widely used
metric to calculate the goodness of the clustering method. We utilize
the silhouette score to evaluate the performance of the metric network.
Specifically, we computed the silhouette score \citep{rousseeuw1987silhouettes}
for each pixel of each image of the validation set. The overall mean
and the class-wise mean silhouette scores of all validation set is
presented in the Table \ref{tab:metric_evaluation}. Note that the
clustering performance correlates with the segmentation performance,
which indicates the collaboration between the metric network and the
segmentation network.

\begin{table}[h]
\centering\caption{Metric Evaluation. The class-wise mean and the overall mean silhouette scores.}

\scalebox{1.0}{%
\begin{tabular}{l|c|c}
Name & Class Mean & Overall Mean\tabularnewline
\hline 
$ST_{MT}$ & 24.3 & 42.0\tabularnewline
$STvM_{Raw}$ & 30.1 & 48.3\tabularnewline
$STvM$ & \textbf{33.5} & \textbf{49.1}\tabularnewline
\hline 
\end{tabular}}

\label{tab:metric_evaluation}
\end{table}

\section{Limitations and Future Work}

The metric network is the essential component of the STvM. It directly
impacts the performance since both the gradient scaling factor and
the patch quality to be used in the MOCM are calculated using the
metric distance. Even though we use highly reliable predictions of
the teacher network, over-confident false predictions may harm the
performance of the metric network. Therefore, better supervision would
be beneficial. Recently, self-supervised training techniques have
shown promising results. Note that metric learning does not need class
labels, but it needs discriminative labels. Therefore, we believe
that training the metric network with self-supervised training techniques
would help the overall performance.

\section{Conclusion}

We proposed a self-training via metric learning (STvM) framework utilizing
the mean-teacher approach for the source-free domain adaptation method
of semantic segmentation. STvM learns a metric feature space directly
in the target domain using a proxy-based metric learning technique.
In training time, a reliability score is calculated for the teacher's
predictions by using the distance of the metric features to the class-proxy
features. The reliability score is used for scaling gradients and
generating object patches. The generated patches are used to augment
the input of the student network with the proposed Metric-based Online
ClassMix (MOCM) method. The experimental results show that utilizing
all the predictions is beneficial for self-training of source-free
domain adaptation and also gradient scaling is useful to mitigate
the negative impact of false positives pseudo-labels. This self-training
strategy also facilitates under-confident positive samples to contribute
to training. MOCM augmentation technique highly perturbs the input
image, facilitating more robust training. STvM significantly outperforms
state-of-the-art methods, and it is highly robust the randomness in
training.

During the preparation of this work the author(s) used ChatGPT in order to improve language and readability. After using this tool/service, the author(s) reviewed and edited the content as needed and take(s) full responsibility for the content of the publication.
\bibliographystyle{model2-names}
\bibliography{references}

\end{document}